# Real-time Yemeni Currency Detection

**Edrees AL-Edreesi**
Computer Science Department,
Faculty of Computer and Information Technology,
Sana'a University, Yemen.
edris.shrf@su.edu.ye

**Ghaleb Al-Gaphari**
Computer Science Department,
Faculty of Computer and Information Technology,
Sana'a University, Yemen.
drghalebh@su.edu.ye

*Abstract*— Banknote recognition is a major problem faced by visually Challenged people. So we propose a application to help the visually Challenged people to identify the different types of Yemenian currencies through deep learning technique. As money has a significant role in daily life for any business transactions, real-time detection and recognition of banknotes become necessary for a person, especially blind or visually impaired, or for a system that sorts the data. This paper presents a real-time Yemeni currency detection system for visually impaired persons. The proposed system exploits the deep learning approach to facilitate the visually impaired people to prosperously recognize banknotes. For real-time recognition, we have deployed the system into a mobile application.

## I. INTRODUCTION

Currently, the use of paper money remains one of the main options for the exchange of products and services. However, one of the remaining problems is the detection of counterfeit banknotes, which increasingly resemble originals, making it difficult for someone who is not an expert in the field to detect them. On the other hand, there are machines for detecting counterfeit banknotes [1]; however, these are often expensive, so the identification and retention of counterfeits ends up falling on financial and government entities, with minimal community involvement [2].

In order to solve this problem and to present alternative solutions, in the state-of-theart, there are proposals based on classical computer vision techniques. For example, from histogram equalization [3], nearest neighbor interpolation [4], genetic algorithms [5] and fuzzy systems [6]. However, the main problem of this type of methods is its low capacity of generalization for new examples as well as its low accuracy. Another group corresponds to those methods based on deep learning (DL) using convolutional neural networks (CNNs) [7–9], which have outperformed to the classic machine learning techniques [10] and humans too [11] in classification tasks.

Considering the current importance of the CNNs in the field of computer vision, there are some proposals in the area of banknote recognition and counterfeit detection. For example, transfer learning (TL) with Histograms of Oriented Gradients for Euro banknotes [12], a YOLO net for Mexican banknotes [13] or custom CNN architectures for dollar, Jordanian dinar and Won Koreano banknotes [14,15] have been proposed. However, one of the main disadvantages of proposals using CNNs that focus on fake banknote recognition is that there is no clarity about which design strategy is more appropriate, either custom or by transfer learning. When using transfer learning-based networks, there are many types of patterns that the network has learned, but they are not specific to the current task. On the other hand, custom networks are trained with a much smaller dataset than the pre-trained networks, but they specifically learn the patterns of this type of classification task. Another shortcoming found in the literature is that the impact of the freezing point (FP) of the pre-trained network on the performance of the classifier has not been analysed.

The main aim of this paper is to introduce an intelligent system that can distinguish between different types of Yemeni paper currencies using deep learning aproach. The system is based on image processing methods to classify different types of currencies efficiently. Moreover, deep learning techniques are used to perform the classification process effectively.

This paper presents a real-time Yemeni currency detection system for visually impaired persons. The proposed system exploits the deep learning approach to facilitate the visually impaired people to prosperously recognize banknotes. For real-time recognition, we have deployed the system into a mobile application.

## II. LITERATURE REVIEW

A large number of researchers have made several contributions toward developing techniques for currency recognition. The different banknote make researchers deal with the recognition task differently for each one of them.

In this section, we review previous work in currency recognition techniques. Mirza and Nanda (Mirza and Nanda, 2012) use three extracted features from the banknote including identification mark, security thread and watermark. The features are extracted using edge based segmentation by Sobel operator. An algorithm based on Local binary patterns (LBP) for recognition of Indian paper currency was proposed by Sharma et al. (2012). The results show that the system provides Currency recognition using a smartphone 485 a good performance for images with low noise with 99% accuracy.

Sargano et al. (2013) proposed a new intelligent system for Pakistani paper currency recognition. The proposed system required less time compared with other systems. Three layer feed-forward Back propagation Neural Network (BPN) is used for classification. The system is tested with 350 Pakistani banknotes. The results indicated that the system had 100% recognition accuracy. This technique is applied on banknotes without any distortion (e.g., wrinkled or folded). Da Costa (da

Costa, 2014) developed a banknote recognition system to recognize multiple banknotes in different perspective views and scales. Feature detection, escription and matching are used to enhance the confidence in the recognition results. The banknote contour is computed using a homography. The system is evaluated with 82 test images, and all the Euro banknotes were successfully recognized. The system provides robust results to handle folded and wrinkled banknotes with several kinds of illumination. The banknote image is converted into gray scale and then compressed. Then each pixel of the compressed image is passed as an input to the network. This method can recognize and match noisy currency images and it provides good results when compared to single network and ensemble network with independent training. The results show the recognition accuracy range from 100% to 54% depending on the noise level of the input image. Vijay and Jain (2013) proposed an image processing technique to extract paper currency denomination. The extr acted Region of Interest (ROI) is used with pattern recognition and neural networks matching technique. In this method they captured the images by the simple flat scanner with a fixed size and then some filters are applied to extract the denomination value of the banknote. The paper has no information about the accuracy of the proposed algorithm.

Reel et al. (2011) use heuristic analysis of characters and digits for Indian currency notes for recognition. This process is invariant to light conditions, use font type and deformations of characters caused by a skew of the image. Heuristic analysis of the characters is performed for this purpose to get the exact features of characters before feature extraction. One of the challenges raised in the character segmentation part is that two characters are sometimes joined together. These techniques focus on extracting the number on the paper currency. However, such technique is not feasible in the case of wrinkled or folded banknote. Paisios et al. (2012) developed a mobile currency recognition system using SIFT to recognize partial images. The system is evaluated using a limited sample set with different state as: folded, incomplete or had orientation and rotation. The results indicated that the nearest neighbor algorithm provides an accuracy of 75% and the nearest to second nearest neighbor ratio algorithm provides an accuracy of 93.83% Toytman and Thambidura (Toytman and Thambidurai, 2011) proposed a banknote recognition system on Android with improved Speed up Robust Features SIFT. The algorithm solves the problems related to illumination conditions, scale and rotation. The approach was insensitive to clutter, occlusion tolerable and wrinkling of banknotes. The approach was tested and good result was obtained with respect to clutter and variations in illumination. On the other hand, the problem of inability to detect folded banknotes is not solved. The paer does not provide any information about the algorithm accuracy.

III. METHODOLOGY

Currency recognition is a field which is studied by many researchers over the last couple of years. Currency can be recognised in a variety of ways depending on features you use for classification as well as machine learning model. Every currency on this planet has certain features on which we can easily classify them. In case of Yemenian currencies, they have some security features which are crucial for recognition of the class. Certain features like the dominant colour of class are one of important features for classification. So, we can infer that a model using grayscale image for currency recognition will surely be left out of one of the important features of classification and can hamper accuracy.

The proposed explanation is reasonably straight forward classifying Yemenian currency notes using deep learning approaches. There are various image classifier models, each following its methodology. The dataset used is images of currency notes segregated into training and testing set by the ratio of 85:15, as shown below. There are approximately 1600 images of different denominations of Yemenian currency.

| Notes | Training | Testing | Total |
|---|---|---|---|
| Hundred | 345 | 55 | 400 |
| Two hundred and fifty | 345 | 55 | 400 |
| Five hundred | 345 | 55 | 400 |
| One thousand | 345 | 55 | 400 |

Fig 1: Distribution of Yemenian Currency Denominations

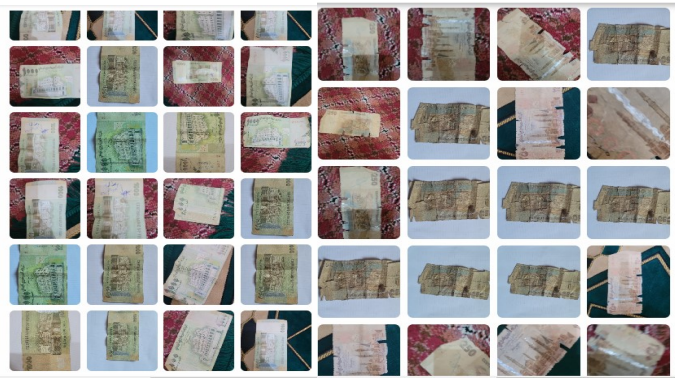

Fig 2: sample from dataset

we used MobleNet as our model for classifying the images. MobileNets are based on a streamlined architecture that utilizes depth-wise detachable convolutions to construct lightweight and compact deep neural networks. We enlist two simple global hyper-parameters that efficiently trade-off between latency and accuracy. These hyper-parameters enable the model builder to select the rightsized model for their application created on the problem's limitations. We shall be using Mobilenet as it is compact in its architecture. It uses depth wise separable convolutions, which implies it conducts a single convolution on each color channel rather than blending all three and flattening it. For MobileNets, the depthwise convolution pertains to a single filter to each input channel. The pointwise convolution then applies a 1×1 convolution to incorporate the outputs of the depthwise convolution. A standard convolution, both filters and combines inputs into a new set of outputs in one step. The depthwise separable convolution slices this into two layers, a separate layer for filtering and a separate layer for combining. This factorization has the consequence of drastically reducing computation and model size.

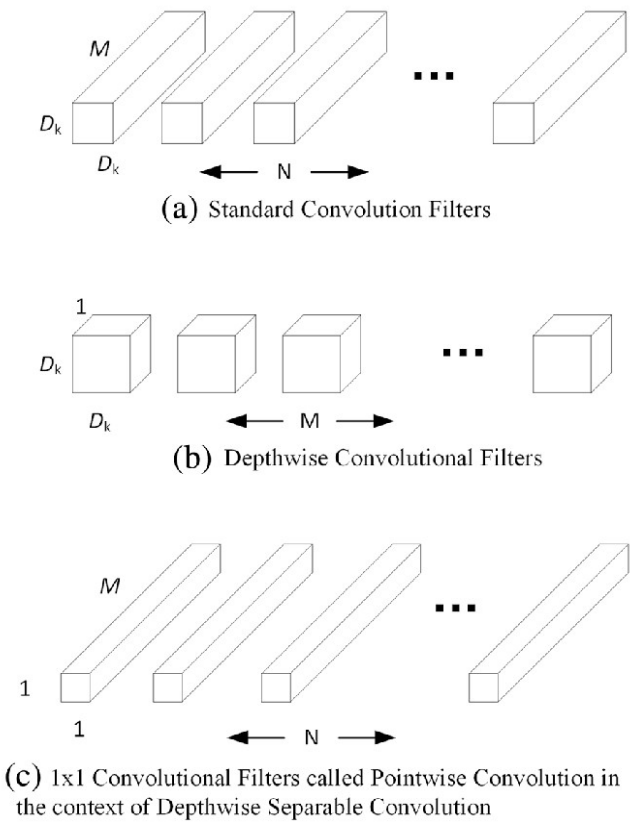

(a) Standard Convolution Filters

(b) Depthwise Convolutional Filters

(c) 1x1 Convolutional Filters called Pointwise Convolution in the context of Depthwise Separable Convolution

The widespread architecture of the Mobilenet is as follows, having 30 layers with
1. convolutional layer with stride 2
2. depthwise layer
3. pointwise layer that doubles the number of channels
4. depthwise layer with stride 2
5. pointwise layer that doubles the number of channels

MobileNet is a neural network architecture that operates extremely efficiently on mobile devices. Its architecture glimpses like as shown in the figure below:

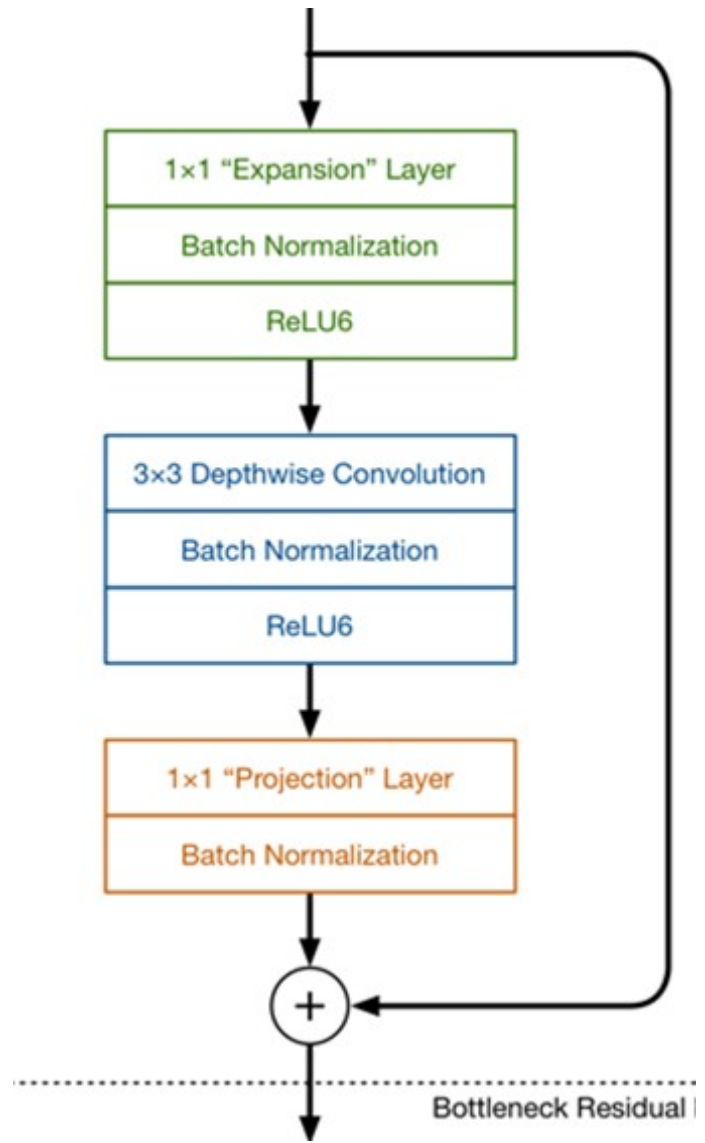

It requires inadequate maintenance, thus accomplishing incredibly competently, delivering high speed. There are several categories of pre-trained models with the network size in memory and disk being proportional to the number of parameters being used. The network's speed and power consumption are proportional to the number of MACs (Multiply-Accumulates), a measure of the number of fused Multiplication and Addition operations

IV. EXPERIMENTS AND RESULTS

The figure below is the representation of the ac3curacy in percentage for different numbers of epochs. We can conclude that from different batch sizes, we got the highest percentage of accuracy in batch size 16.

**Accuracy Per class**

For this model it is necessary to check the accuracy of every denomination to check the currency note correctly. As on the teachable machine we created the model of different classes containing the denominations, and then setting proper epoch value the below accuracy achieved. For this image processing project, we included all the Yemenian denominations ranging from 100 to 1000.

Accuracy per class

| CLASS | ACCURACY | # SAMPLES |
|-------|----------|-----------|
| 1000  | 1.00     | 29        |
| 250   | 1.00     | 21        |
| 100   | 1.00     | 18        |

**Accuracy per epoch**

Also, the loss factor is considered to achieve better accuracy through epoch. For this the data set is trained for the 50 epoch and then graphical accuracy is achieved for images included in the dataset.

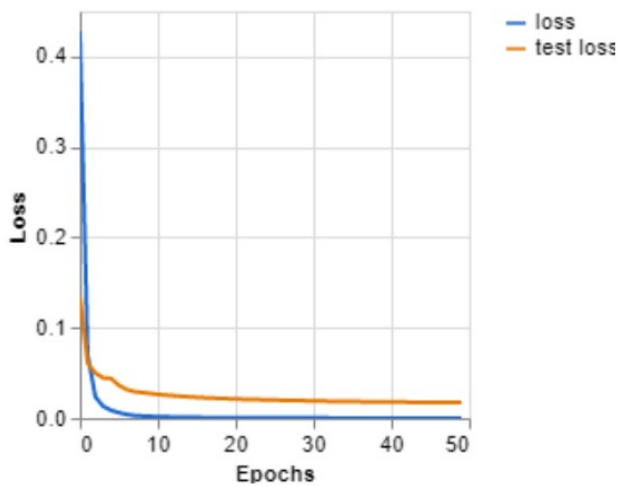

**Accuracy on Teachable Machine**

When we capture any image of a currency note then the model identifies the correct denomination through the trained dataset. As for e.g., the denomination of 1000 is captured by the camera module then it is checked with the trained data with the help of tflite and as we can see through the output table the accuracy for the 1000- currency note is 100% it means technique works properly. The same happens with the other currency denominations.

The deployment of the android application is shown below in 3 steps, same as a web application:

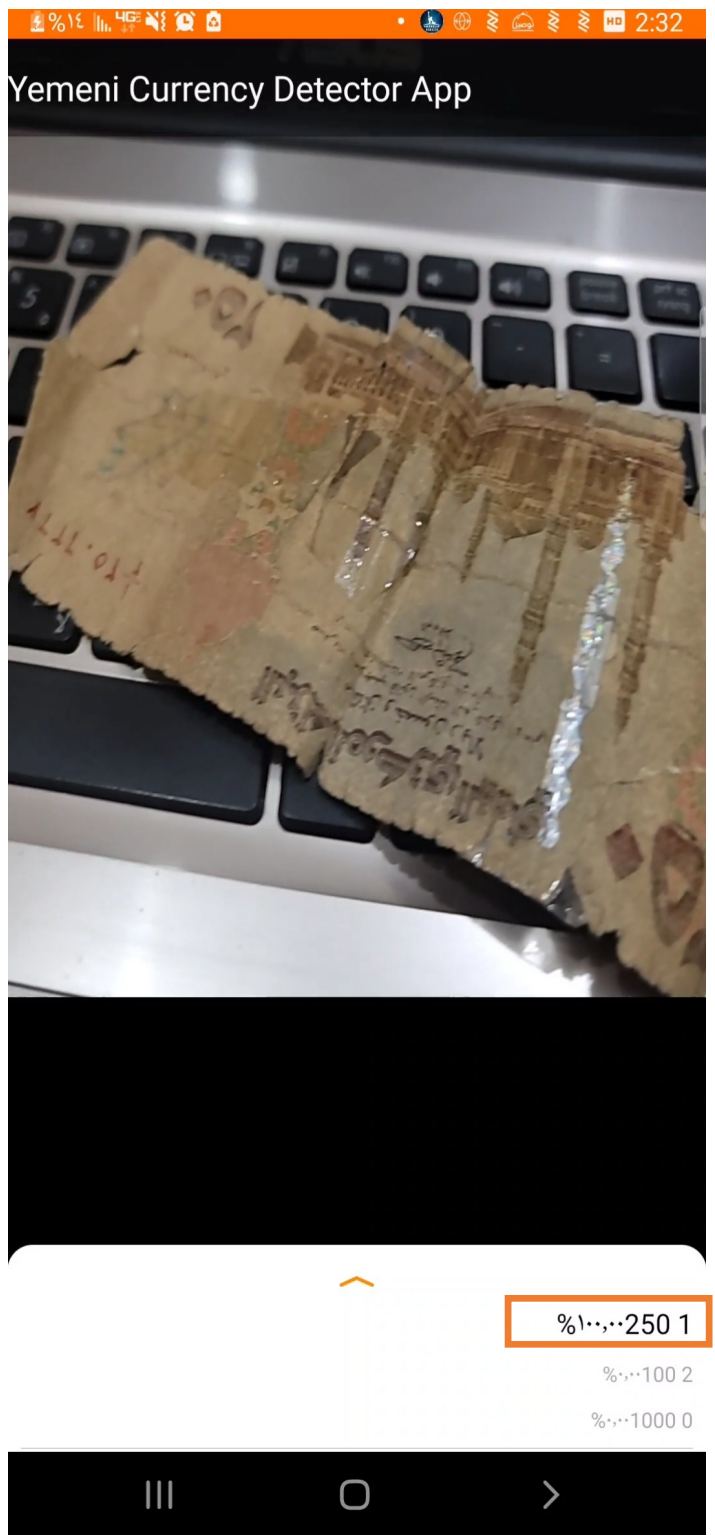

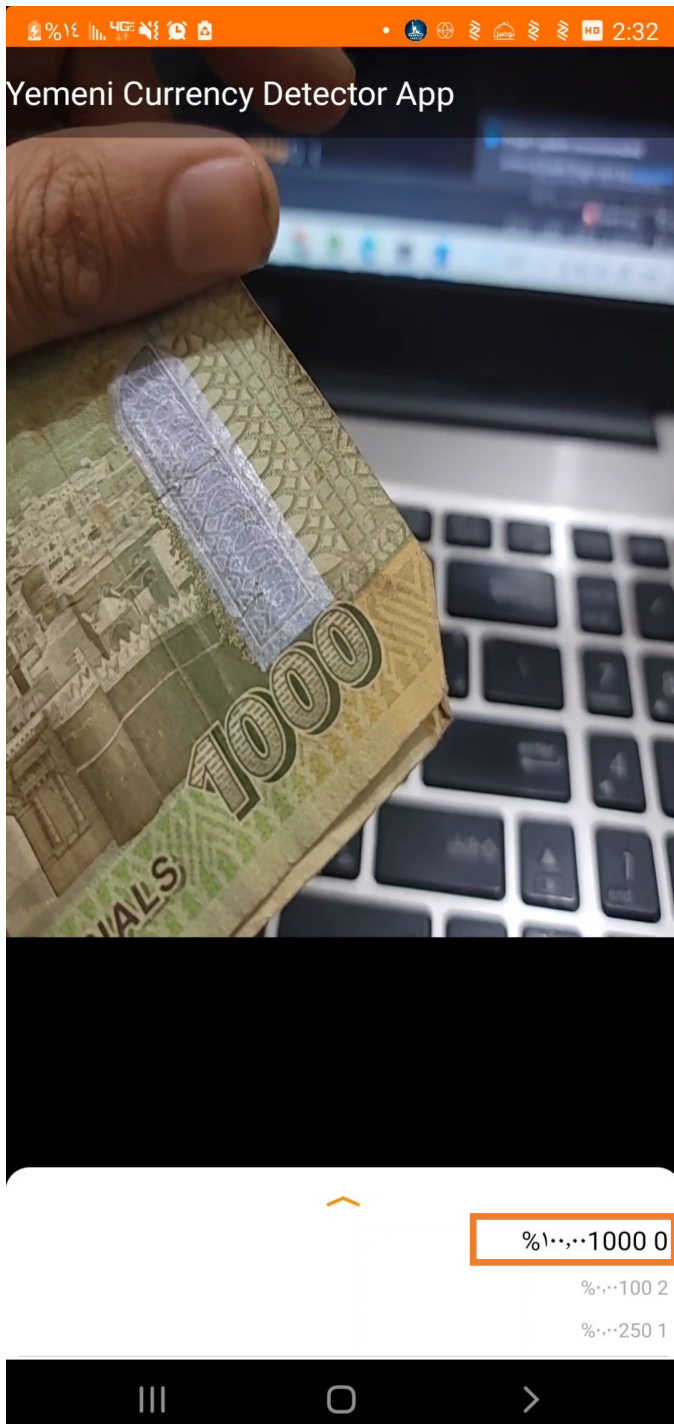

## V. CONCLUSION

In this paper, we have explained the model we constructed to classify Indian currency notes. We have also elucidated the notion behind this, i.e., convolution neural network (CNN) algorithm and transfer learning; which helped extract features of the currency note. Out of all the models we tested, MobileNet bestowed us with the highest accuracy for our dataset. We then deployed the model with an android app using Android developed by google to make it serviceable for the user. Our current restrictions include a lack of risk analysis. Due to given time constraints, we aim to optimize and better accuracy in the next phase. In the future, we will improve on a better dataset, which in turn provides hereafter giving much better accuracy and results. We can also aim to work on the UI part of the android app to make it more understandable for the end-user in the future.

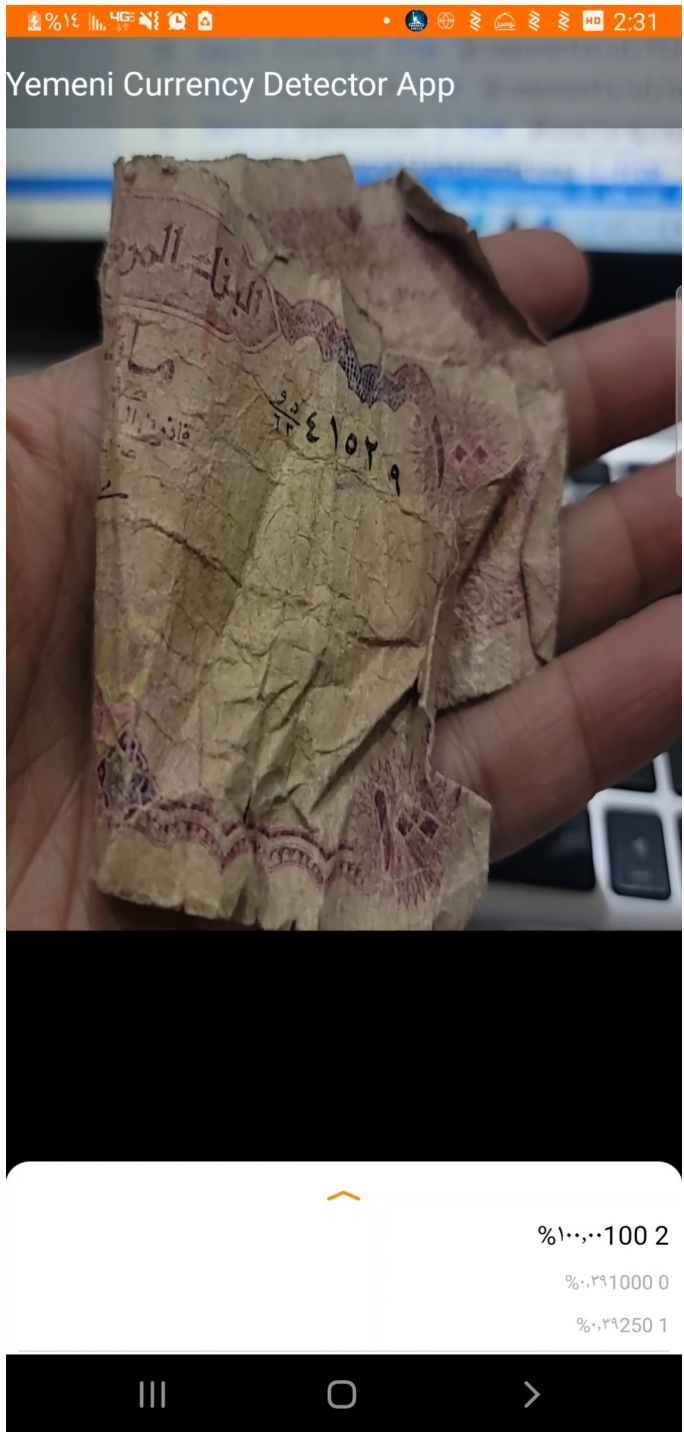
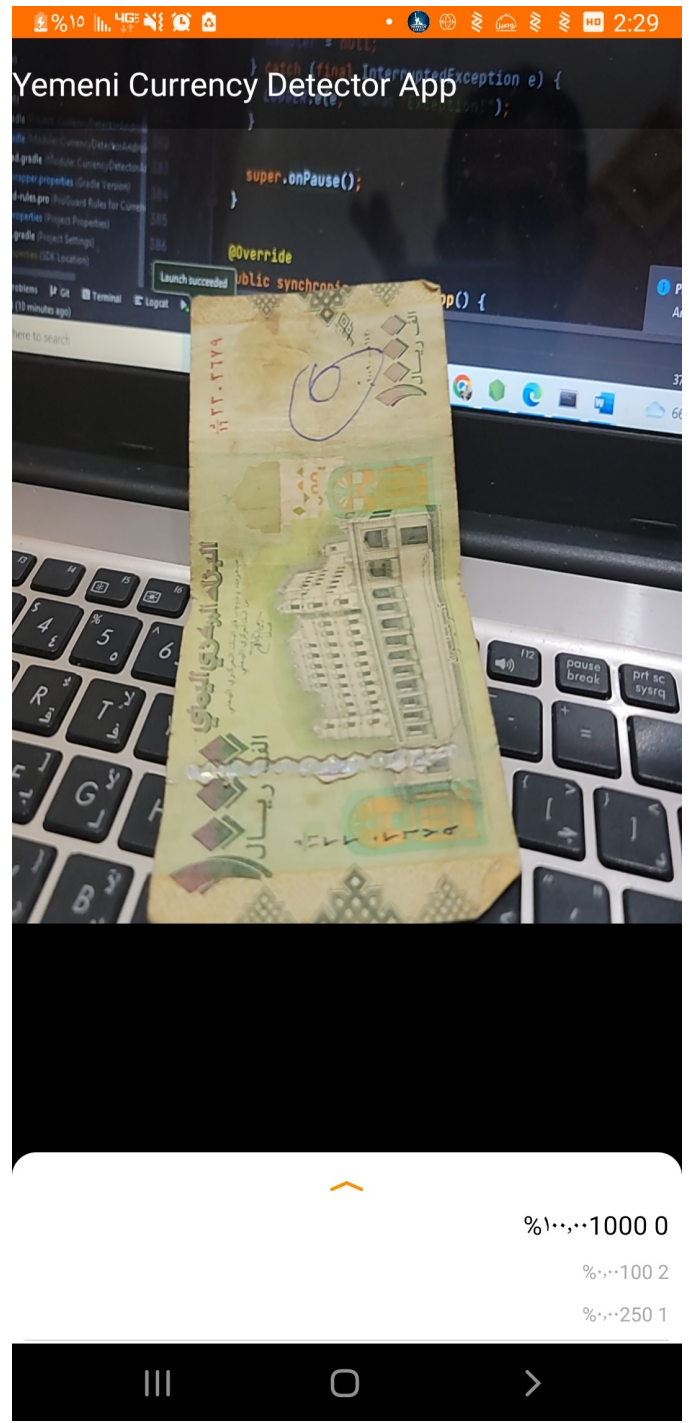